\documentstyle[psfig]{llncs}

\newcommand{\comment}[1]{}
\newcommand{\myspace}{\mbox{ }\ \ \ \ }

\newcommand{\sep}{\,\mid\,}
\newcommand{\mydot}{\raisebox{.15em}{\tiny $\,\bullet\,$}}
\newcommand{\size}[1]{\left | {#1} \right |} 
 
\newcommand{\order}[1]{{\cal O}(#1)}

\newcommand{\cfg}{CFG}

\newcommand{\nonterm}{\mbox{$V_{{\rm N}}$}}
\newcommand{\myterm}{\mbox{$V_{{\rm T}}$}}

\newcommand{\de}{\rightarrow}
\newcommand{\derivg}[1]{\mathrel{\mbox{$\:\Rightarrow\:$}}}
\newcommand{\derivrg}[2]{\mathrel{\mbox{$\:\stackrel{\!{#1}}%
        {\Rightarrow\!}\:$}}}

\newcommand{\ep}{\varepsilon}

\newcommand{\eset}{I_{{\rm E}}}
\newcommand{\ieset}{I_{{\rm V}}}
\newcommand{\chartset}{I_{{\rm C}}}

\newtheorem{alg}{Algorithm}

\begin{document}

\title{A Variant of Earley Parsing}

\author{Mark-Jan Nederhof\inst{1}%
	\thanks{Research by the first author is carried out
		within the framework of the Priority Programme
		Language and Speech Technology (TST). The TST-Programme is
		sponsored by NWO (Dutch Organization for Scientific 
		Research).} 
      and Giorgio Satta\inst{2}}

\institute{Faculty of Arts \\
		University of Groningen \\
		P.O. Box 716 \\
		NL-9700 AS Groningen \\
		The Netherlands \\
                {\tt markjan@let.rug.nl} 
		\and
        	Dipartimento di Elettronica ed Informatica \\ 
        	Universit\`{a} di Padova \\
        	via Gradenigo, 6/A \\
        	I-35131 Padova \\
		Italy \\
                {\tt satta@dei.unipd.it}}

\maketitle

\begin{abstract}
The Earley algorithm is a widely used parsing method in 
natural language processing applications. 
We introduce a variant of Earley parsing that is based on
a ``delayed'' recognition of constituents.  This allows us 
to start the recognition of a constituent only in cases in 
which all of its subconstituents have been found within the 
input string.  This is particularly advantageous in several 
cases in which partial analysis of a constituent cannot be 
completed and in general in all cases of productions sharing 
some suffix of their right-hand sides (even for different 
left-hand side nonterminals). 
Although the two algorithms result in the same asymptotic
time and space complexity, from a practical perspective our 
algorithm improves the time and space requirements of 
the original method, as shown by reported experimental 
results. 
\end{abstract}

\section{Introduction}

Earley parsing is one of the most commonly used methods
for the (automatic) syntactic analysis of natural language sentences,
given a context-free grammar model.  
This method does not use backtracking, resulting in time 
and space efficiency, and is quite flexible, in that it does
not require the input grammar to be cast in any particular form.
Earley parsing was first defined 
in~\cite{Earley:70}, in the context of formal language parsing.
This method has later been rediscovered in~\cite{Kaplan:73,Kay:86a} 
from the perspective of application to natural language processing, 
where it was called {\em active chart parsing}.  Active chart parsing
makes also use of a data structure, called {\em agenda}, which allows a
more flexible control of competing analyses.  

A considerable number of results and applications regarding Earley parsing
have been published in the literature. 
{}From a theoretical perspective, improvements of the Earley algorithm 
have been reported in~\cite{Graham:80}, \cite{Leiss:90} and~\cite{LE91b}.  
Several reformulations
of Earley parsing have also been presented.  Most remarkably, 
in~\cite{Billot:89} Earley parsing is related to 
the deterministic simulation of
a particular kind of nondeterministic pushdown automaton, and 
a recursive reformulation of Earley parsing has been 
proposed in~\cite{Leermakers:92}. 

{}From the perspective of natural language parsing, 
the Earley method has been adapted to work with context-free grammars
enriched with feature structures in~\cite{PE83}, \cite{Shieber85AC}
and~\cite{Gardemann:89}, 
and to cope with on-line semantic interpretation in~\cite{Stock:89CL}. 
Comparison of Earley parsing with other parsing strategies has 
been experimentally carried out and reported  
in~\cite{Wiren:87EA} and~\cite{Shann:91}.

In this paper we focus on a drawback of the
Earley algorithm: the recognition of a production within the input
is started by looking for the constituents in its right-hand side, 
proceeding from left to right.  
In this process, the algorithm keeps track of the position 
within the input at which the recognition has started.  
Since this information is needed only if the whole recognition
can be carried to an end, the algorithm behaves 
in a rather inefficient way in several cases in which production 
recognition cannot be successfully completed. 
We propose a variant of the original method, in which 
the problem is solved by delaying some of the computation 
until the involved productions have been fully recognized. 
This is achieved using an idea first presented 
in~\cite{LE92} in the context of left-corner parsing, 
as it will be discussed at length in the final section.
When applied in the framework of active chart parsing, 
our technique results in the ``inversion'' of the fundamental 
rule~\cite{Kaplan:73,Kay:86a} that combines a left active 
edge with a right inactive edge. 
Although our proposal does not result in an asymptotic
improvement of the time and space complexity of the Earley algorithm, 
reported experimental results provide evidence that in practical cases 
our method achieves an increase in time and space efficiency.

The remainder of this paper is organized as follows.
In Section~\ref{s:prel} some preliminaries are discussed.
We review the Earley parsing method in Section~\ref{s:earley},
and then introduce our variant in Section~\ref{s:variant}.
Some empirical results are given in Section~\ref{s:empiric},
and related work is discussed in Section~\ref{s:disc}.

\section{Preliminaries}
\label{s:prel}

We introduce the formal notation that will be used throughout the paper.  

A string $w$ is a finite sequence of symbols over some alphabet.
We denote as $\size{w}$ the length of $w$, and 
as $\ep$ the (unique) string of length zero. 
The set of all strings over some alphabet $\Sigma$, $\ep$ included, 
is denoted $\Sigma^\ast$. 
A context-free grammar (\cfg) is a rewriting system 
$G = (\myterm, \nonterm, P, S)$, where $\myterm$ and $\nonterm$
are two finite, disjoint sets of terminal 
and nonterminal symbols, respectively, $S \in \nonterm$ is the start symbol, 
and $P$ is a finite set of productions. Each production has the form
$A \de \alpha$ with $A \in \nonterm$ and 
$\alpha \in (\nonterm \cup \myterm)^\ast$. 
The size of $G$, written $\size{G}$, is defined
as $\sum_{(A \de \alpha)\in P} \size{A\alpha}$.

We generally use symbols $A, B, C, \ldots$ to range over $\nonterm$,
symbols $a,b,c, \ldots$ to range over $\myterm$,
symbols $X, Y$ to range over $\nonterm \cup \myterm$,
symbols $\alpha, \beta, \gamma, \ldots$ to range over 
$(\nonterm \cup \myterm)^\ast$,
and symbols $v, w, x, \ldots$ to range over $\myterm^\ast$.
For a fixed grammar,
the binary relation $\derivg{G}$ is defined over $(\nonterm \cup \myterm)^\ast$
such that $\gamma A \delta \derivg{G} \gamma \alpha \delta$ whenever
$A \de \alpha$ belongs to $P$.  
We will mainly use the reflexive 
and transitive closure of $\derivg{G}$, denoted $\derivrg{\ast}{G}$. 

\section{Earley Parsing}
\label{s:earley}

We briefly present here the Earley algorithm,
before introducing the variant of this method in the next section.

Let $G = (\myterm, \nonterm, P, S)$ be a \cfg. 
We associate with $G$ a set of symbols, 
called {\em dotted items}, specified as: 
\begin{eqnarray}
\eset & = & \{ [A \de \alpha \mydot \beta] \sep 
	(A \de \alpha \beta) \in P \}.
\end{eqnarray} 
Dotted items are used below to represent intermediate steps in
the process of recognition of a production of the grammar, 
where the sequence of symbols in between the arrow and the dot 
indicates the sequence of constituents recognized so far
at consecutive positions within the input string. 
More precisely, given a production $p: (A \de X_1 X_2 \cdots X_r)$,
$r \geq 0$, the process of recognition of the right-hand side of
$p$ is carried out in several steps. 
We start from item $A \de \mydot X_1 X_2 \cdots X_r$, 
attesting that the empty sequence of constituents has been 
collected so far.  
This item represents a prediction for $p$. 
We then proceed with item $A \de X_1 \mydot X_2 \cdots X_r$
after the recognition of a constituent $X_1$, and so on.  
Production $p$
has been fully recognized only if we reach item 
$A \de X_1 X_2 \cdots X_r \mydot$, attesting therefore the 
complete recognition of a constituent $A$. 
In active chart parsing, items in $\eset$ with the dot 
not at the rightmost position of the right-hand side 
are used to label the so called {\em active edges}.

Given a string $w = a_1 a_2 \cdots a_n$, with 
$n \geq 0$ and each $a_i$ a terminal symbol,
we call {\em position\/} within $w$ any integer $i$ 
such that $0 \leq i \leq n$. 
In what follows, $E$ is a square matrix whose entries are 
subsets of $\eset$ and are addressed by indices that are 
positions within the input string. Entries are denoted as $E_{i,j}$. 
The insertion by the algorithm of item 
$[A \de \alpha \mydot \beta]$ in $E_{i,j}$, $i \leq j$, 
attests the fact that the sequence of constituents in $\alpha$
exactly spans the substring $a_{i+1} \cdots a_j$ of the input.
(See below for a more precise characterization of the algorithm.)
Control flow is not specified in the method below, since it
is usually regulated by means of a data structure called {\em agenda}, 
which directs the incremental construction of the table by means of an
iteration: starting from an empty table, items are added as long as needed,
and with the desired priority.

\begin{alg}[Earley] \rm
\label{a:earley}
Let $G = (\myterm, \nonterm, P, S)$ be a \cfg.
Let $w = a_1 a_2 \cdots a_n$ be an input string, $n \geq 0$, 
and $a_i \in \myterm$ for $1 \leq i \leq n$. 
Compute the least $(n+1) \times (n+1)$ 
table $E$ such that $[S \de \mydot \alpha] \in E_{0,0}$ 
for each $(S \de \alpha) \in P$, and 
\\[.5ex]
\begin{tabular}{lll}
1.\hspace{0.5em} & $[A \de \mydot \gamma] \in E_{j,j}$ &
        if $[B \de \alpha \mydot A \beta] \in E_{i,j}$,
                    $(A \de \gamma) \in P$;
\\
2. & $[A \de \alpha a_j \mydot \beta] \in E_{i,j}$\ \ \ \ \  &
        if $[A \de \alpha \mydot a_j \beta] \in E_{i,j-1}$;
\\
3. & $[A \de \alpha B \mydot \beta] \in E_{i,j}$ &
        if $[A \de \alpha \mydot B \beta] \in E_{i,k}$,
        $[B \de \gamma \mydot] \in E_{k,j}$.
\end{tabular}
\\[.5ex]
The string $w$ is accepted if and only if 
$[S \de \alpha \mydot] \in E_{0,n}$ for some $(S \de \alpha) \in P$.
\end{alg}
The correctness of the algorithm immediately follows from the
property below, whose proof can be found 
in~\cite{Earley:70} and~\cite{Graham:76}. 

\begin{proposition}
In Algorithm~{\/\rm \ref{a:earley}}, an item $[A \de \alpha \mydot \beta]$
is inserted in $E_{i,j}$ if and only if the following 
conditions hold:
\\[-1.5em]
\begin{description}
\item[{\it A1.}] 	$S \derivrg{\ast}{G} a_1 \cdots a_i A \gamma$, some 
        $\gamma$;  and 
\item[{\it A2.}]   $\alpha\derivrg{\ast}{G} a_{i+1} \cdots a_{j}$.
\end{description}
\label{p:earley}
\end{proposition} 
\vspace{-2.2ex}
For methods cruder than the Earley algorithm, 
membership of an item in some entry may merely be subject to
condition~{\it A2}, which is sufficient for determining the 
correctness of the input. 
However, Earley's algorithm is more selective, as is apparent from
condition~{\it A1}, which characterizes the 
so called top-down filtering capability of the method. 
Condition~{\it A1} guarantees that only those constituents are predicted
that are compatible with the portion of the input
that has been read so far.

Assuming the working grammar as fixed, 
a simple analysis reveals that Algorithm~\ref{a:earley} runs in 
time $\order{n^3}$.\footnote{%
When both the input string and the grammar are taken as input parameters,
Algorithm~\ref{a:earley} runs in time $\order{\size{G}^2n^3}$.
An improvement of Algorithm~\ref{a:earley} 
has been presented in~\cite{Graham:80}, running in time 
$\order{\size{G}n^3}$.} 
This will be more carefully discussed in the next section.

\section{A Variant of Earley Parsing}
\label{s:variant}

In this section we introduce a variant of Earley parsing
that can be obtained by reconsidering the way in which 
the results of the intermediate steps are stored in the process of production 
recognition. 

Let us focus on the dependence of the running time of Algorithm~\ref{a:earley}
on the length of the input string. From this perspective, the most expensive 
step is Step~3.  Intuitively, this is the case because there 
might be $\order{n^2}$ items that are inserted at this step in some entry 
of $E$, and each item can in turn be the result of $\order{n}$ 
different combinations of pairs of items already in $E$. 
In practice, the total number of different combinations of dotted items
attempted by Step~3 when processing an input string
dominates the running time of Algorithm~\ref{a:earley}. 
The change to the new method consists in
a decomposition of Step~3
that results, in some cases, in a reduction of this number. 
We introduce the basic idea through an example.

\begin{figure}[ht]
\centerline{\psfig{figure=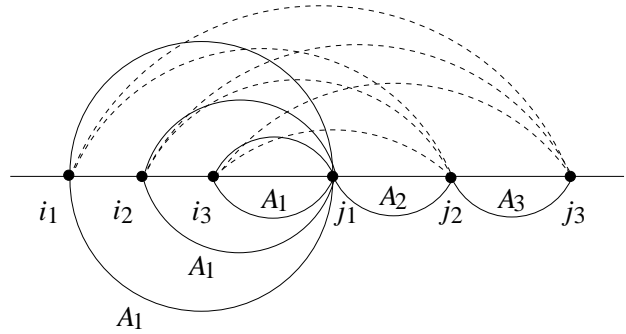}} 
\caption{We depict the case of $d = 3$, $r = 4$, and assume 
$D = \{i_1, i_2, i_3\}$.  We represent the input string by means of
an horizontal line and each dotted item in $E$ by means of an arc; 
only the relevant positions within the input string are depicted.
In the attempt to recognize production $A \de A_1 \cdots A_4$,
the algorithm has created $3$ dotted items
$[A \de A_1 \protect\mydot A_2 A_3 A_4]$, 
one for each position in $D$, 
depicted by solid arcs above the horizontal line. 
Since each of these items has a different left position, 
the Earley algorithm is forced to instantiate $3$ independent 
processes for the recognition of $A \de A_1 \cdots A_4$. 
These processes will create the dotted items depicted by 
the dashed arcs. Note that in collecting the remaining 
constituents $A_2, A_3, A_4$ the method duplicates the needed effort.}
\label{f:earley}
\end{figure}

Consider a production $p: (A \de A_1 A_2 \cdots A_r)$, $r \geq 3$.
Let $D$ be a set containing $d > 2$ positions within the input string. 
Assume that the dotted item $[A \de A_1 \mydot A_2 \cdots A_r]$
has been inserted in the entry $E_{i,j_1}$, for each $i \in D$
and for some fixed $j_1$. 
This corresponds to $d$ constituents $A_1$ recognized within the input. 
Assume also that, for each $t$ with $2 \leq t \leq r-1$, a constituent $A_t$
has been recognized in entry $E_{j_{t-1},j_t}$. 
Finally, assume that no constituent $A_r$ is found starting
at position $j_{r-1}$ (see Figure~\ref{f:earley}). 
Under these assumptions, Step~3 will be executed 
$d(r-2)$ more times, carrying out $d$ independent recognition processes
for $p$, to find out at the end that none of these processes can be 
successfully completed, because of the lack of constituent $A_r$. 
The fact that the above recognition processes are independent one
of the other is due to the fact that in Step~3
we record the position within the input where each 
process started (the positions in $D$). 

We observe that the left position of $p$ in the input string is needed only 
if the recognition process of $p$ can be successfully completed, 
in order to locate the constituent corresponding to 
the left-hand side of $p$ for use in the remaining analysis of the input.  
On this basis, we reformulate Step~3 by splitting it up into
two substeps. 
The first substep performs the recognition of $p$ in a forward
manner, without maintaining any record of the left position. 
This is done using an array $U$ in whose entries we store 
only the suffixes of $p$'s right-hand side that must still be recognized. 
If the recognition can be successfully completed, we apply
the second substep and compute the left positions of $p$ in a 
backward manner, 
starting from the rightmost constituent in $p$'s right-hand side 
and proceeding toward the left, 
storing the intermediate results in table $T$.  

The proposed technique thus delays part of the computation
from the former Step~3 until we are granted that $p$ can
be successfully recognized.  In this way we avoid the 
computational inefficiency revealed by our example. 
In fact, whenever $p$'s recognition cannot 
be completed, no backward computation is performed by 
our method, resulting in some time and space savings. 
More precisely, the same computation performed by 
the $d(r-2)$ executions of Step~3 in 
Algorithm~\ref{a:earley} will be performed by 
$r-2$ executions of the forward substep, and $0$ executions
of the backward substep.
In addition we observe that, even in the presence of 
a constituent $A_r$ with left position $j_{r-1}$ in the 
input string, the proposed technique performs more 
efficiently than the original formulation of Step~3.
In fact, since the backward substep proceeds 
from right to left, constituents $A_r, A_{r-1}, \ldots$
will be visited only once in the attempt to find all possible 
left positions for $p$.   

We observe that for the technique described above to work in its full
generality, also 
Step~2 from Algorithm~\ref{a:earley} should be split into two substeps.
This allows correct treatment of productions containing terminal symbols 
in their right-hand sides.
Finally, it is not difficult to see that the problem described above 
can be generalized to productions sharing some suffix 
of their right-hand sides, that is productions of the form 
$A \de \alpha \gamma$ and $B \de \beta \gamma$, 
in cases that $\gamma$ is, at some position, predicted independently 
for both productions.

We are now in a position to give a precise specification of
the proposed parsing algorithm. 
Let $G = (\myterm, \nonterm, P, S)$ be a \cfg. 
We associate with $G$ a set of symbols, 
called {\em suffix items}, specified as: 
\begin{eqnarray}
\label{e:ieset}
\ieset & = & \{ [\beta] \sep (A \de \alpha \beta) \in P \}.
\end{eqnarray}
Suffix items serve two different purposes.
First, the insertion of suffix item $[\alpha]$ 
in entry $U_{j}$, where $U$ is a one-dimensional array, 
means that the process of forward recognition
of a production $A \de \alpha \beta$, for some $A$ and $\alpha$, 
has been successfully carried out, up to position $j$ and 
up to the constituents in the sequence $\alpha$.  
In other words, there exists at least one $i$, $i \leq j$, such that some 
dotted item $[A \de \alpha \mydot \beta]$ would have 
been inserted in $E_{i,j}$ by Algorithm~\ref{a:earley}. 
Second, the insertion of suffix item $[\beta]$ in $T_{i,j}$ means that 
at least one production $A \de \alpha \beta$, for some $A$ and $\alpha$, 
has been completely recognized
and the constituents in the sequence $\beta$ have been 
collected backwards
so far, spanning the substring $a_{i+1} \cdots a_j$.

\begin{alg}[Variant of Earley] \rm
\label{a:variant}
Let $G = (\myterm, \nonterm, P, S)$ be a \cfg.
Let $w = a_1 a_2 \cdots a_n$ be an input string, $n \geq 0$,  
and $a_i \in \myterm$ for $1 \leq i \leq n$. 
Compute the least $(n+1) \times (n+1)$ 
table $T$ and the least $(n+1)$ array $U$
such that $[\alpha] \in U_{0}$ 
for each $(S \de \alpha) \in P$, and 
\\[.5ex]
\begin{tabular}{lll}
1.\hspace{0.5em} & $[\gamma] \in U_{j}$ &
        \myspace if $[A \beta] \in U_{j}$,
        $(A \de \gamma) \in P$;
\\
2. & $[\beta] \in U_{j}$ &
        \myspace if $[a_j \beta] \in U_{j-1}$;
\\
3. & $[\beta] \in U_{j}$ &
        \myspace if $[B \beta] \in U_{k}$,
        $(B \de \gamma) \in P$,
        $[\gamma] \in T_{k,j}$;
\\
4. & $[\ep] \in T_{m,m}$ &
        \myspace if $[\ep] \in U_{m}$;
\\
5. & $[a_{j} \beta] \in T_{j-1,m}$ &
        \myspace if $[a_{j} \beta] \in U_{j-1}$,
        $[\beta] \in T_{j,m}$;
\\
6. & $[B \beta] \in T_{k,m}$ &
        \myspace if $[B \beta] \in U_{k}$,
        $(B \de \gamma) \in P$,
        $[\gamma] \in T_{k,j}$,
        $[\beta] \in T_{j,m}$.
\end{tabular}
\\[.5ex]
The string $w$ is accepted if and only if 
$[\alpha] \in T_{0,n}$ for some $(S \de \alpha) \in P$. 
\end{alg}
Step~1 of Algorithm~\ref{a:earley} exactly corresponds to 
Step~1 of Algorithm~\ref{a:variant}.  
Step~2 of Algorithm~\ref{a:earley} has now been split into 
Steps~2 and~5 of Algorithm~\ref{a:variant}, 
which act as forward and backward substeps, respectively. 
Similarly, Step~3 of Algorithm~\ref{a:earley} has been split into 
Steps~3 and~6 of Algorithm~\ref{a:variant}. 
Step~4 of Algorithm~\ref{a:variant} 
is needed to initiate the backward process of
recognizing a production, after the forward process has completed
recognition of the right-hand side.

The correctness of the method directly follows from the 
property stated below, which characterizes the presence of 
suffix items in entries of $U$ and $T$.  

\begin{proposition}
In Algorithm~{\/\rm \ref{a:variant}}, an item $[\beta]$
is inserted in $U_{j}$ if and only if
the following
conditions hold:
\\[-1.5em]
\begin{description}
\item[{\it A1.}]   $S \derivrg{\ast}{G} a_1 \cdots a_i A \gamma$, 
        some $i$, $A$ and $\gamma$;  
\item[{\it A2.}]    $(A \de \alpha \beta)\in P$, some 
        $\alpha$; and
\item[{\it A3.}]   $\alpha\derivrg{\ast}{G} a_{i+1} \cdots a_{j}$,
\end{description}
\vspace{-.7ex}
and an item $[\beta]$ is inserted in $T_{j,k}$ if and only if the following
conditions hold:
\\[-1.5em]
\begin{description}
\item[{\it B1.}] the conditions {\it A1}, {\it A2} and {\it A3} hold; and
\item[{\it B2.}] $\beta\derivrg{\ast}{G} a_{j+1} \cdots a_{k}$.
\end{description}
\label{p:variant}
\end{proposition}
\vspace{-2.2ex}
The proof of the above statement is similar to that of 
Proposition~\ref{p:earley}.

It is not difficult to see that Algorithm~\ref{a:variant}
has running time $\order{n^3}$
(again, we assume the working grammar is fixed). 
Therefore Algorithms~\ref{a:earley} and~\ref{a:variant}
present the same asymptotic time complexity. 
For the purpose of more carefully
comparing the two algorithms, we give below an alternative
to Proposition~\ref{p:variant}, which characterizes the entries in $U$ and $T$ 
in terms of the entries in $E$.

\begin{proposition}
In Algorithm~{\/\em \ref{a:variant}}, an item $[\beta]$
is inserted in $U_{j}$ if and only if
the following
condition holds:
\\[-1.5em]
\begin{description}
\item[{\it A1.}] at least one item $[A\de\alpha\mydot\beta]$ is inserted in $E_{i,j}$
by Algorithm~{\/\em \ref{a:earley}}, for
some $A$, $\alpha$ and $i$, 
\end{description}
\vspace{-.7ex}
and an item $[\beta]$ is inserted in $T_{j,k}$ if and only if the following
conditions hold:
\\[-1.5em]
\begin{description}
\item[{\it B1.}]  the condition {\it A1} holds; and
\item[{\it B2.}]  $\beta\derivrg{\ast}{G} a_{j+1} \cdots a_{k}$.
\end{description}
\label{p:dependent}
\end{proposition}
\vspace{-2.2ex}
This proposition clearly shows that the number of items in $U$ is always
smaller than the number of items in $E$: several items 
$[A\de\alpha\mydot\beta]$ in $E_{i,j}$ for fixed $j$ but differing
$A$, $\alpha$ and $i$ correspond to one single item $[\beta]$ in $U_j$.

On the other hand, the number of items in $T$ may be larger than the number of 
items in $E$ since for each $[A\de\alpha\mydot\beta]$ in $E_{i,j}$ we may have
$[\beta]$ in several $T_{j,k}$ for distinct values of $k$. Since there may be up to $n$ such $k$ in the worst case, the number of items in $T$ may 
be up to $n$ times larger than the number of items in $E$. 

One example of a \cfg\ were this phenomenon is apparent is the following. 
\begin{quote}
\hspace*{\fill} 
$S \de {\it AB}$ \hfill
$A \de {\it C}$ \hfill
$B \de {\it C}$ \hfill
$C \de {\it aC}$ \hfill
$C \de \ep$
\hspace*{\fill}
\end{quote}
For input $a^n$, some $n$,
Algorithm~\ref{a:earley} computes $n+1$ items of the
form $[S\de A \mydot B] \in E_{0,i}$, $0\leq i \leq n$, and $n+1$ items
of the form $[S\de {\it AB} \mydot] \in E_{0,j}$, $0\leq j \leq n$.
On the other hand, Algorithm~\ref{a:variant} computes 
$\frac{n^2+n}{2}$
items of the form $[B] \in T_{i,j}$, $0\leq i \leq j \leq n$.

We define $|E|=\Sigma_{i,j} |E_{i,j}|$,
$|U|=\Sigma_{i} |U_{i}|$, $|T|=\Sigma_{i,j} |T_{i,j}|$, and summarize the
above as follows.
\begin{proposition}
For a fixed \cfg\ and input of length $n$,
let $E$ be constructed by Algorithm~{\/\rm \ref{a:earley}} and 
$U$ and $T$ by Algorithm~{\/\rm \ref{a:variant}}.
Then:
\\[-1.5em]
\begin{enumerate}
\item $|U| \leq |E|$; and
\item $|T| \leq n \cdot |E|$.
\end{enumerate}
\label{p:tabsize}
\end{proposition}
\vspace{-2.2ex}
The second part of this proposition seems to suggest that the table size may be
much larger for the variant. The empirical data presented by
the next section however show that such worst-case behaviour does not seem to 
occur for the practical grammars at hand.

Based on the number of items that are stored in the respective tables,
we can investigate the number of steps that are performed by the two
algorithms. 
We count the number of elementary parsing steps consisting
in the derivation of one item in a table
from one or more objects, such as productions, input symbols, or
other items in a table.
For example, in the case of Algorithm~\ref{a:variant} every combination of
four objects of the form
$[B \beta] \in U_{k}$, $(B \de \gamma) \in P$, $[\gamma] \in T_{k,j}$, and
$[\beta] \in T_{j,m}$ is counted as one elementary parsing
step according to Step~6. 
For a certain \cfg\ and input,
let us denote the number of applications of Steps~1, 2 and~3
of the Earley algorithm 
by ${\cal E}_1$, ${\cal E}_2$ and ${\cal E}_3$. 
Similarly, we introduce the notation
${\cal V}_1$, \ldots, ${\cal V}_6$ for the six steps of the variant.
We further define ${\cal E}={\cal E}_1+{\cal E}_2+{\cal E}_3+
\size{\{\alpha\ |\ (S \de \alpha) \in P\}}$, and
${\cal V}={\cal V}_1+{\cal V}_2+\cdots+{\cal V}_6+
\size{\{\alpha\ |\ (S \de \alpha) \in P\}}$.

Based on condition~{\it A1\/} in Proposition~\ref{p:dependent}, 
we may conclude
that ${\cal V}_1 \leq {\cal E}_1$, ${\cal V}_2 \leq {\cal E}_2$
and ${\cal V}_3 \leq {\cal E}_3$. 
The number of applications of Step~4 is bounded by the number of items
$[\ep]\in U_{j}$, which is bounded by the number of items 
$[A\de\gamma\mydot]\in E_{i,j}$. 
This in turn is bounded by the number of
items $[A\de\mydot\gamma]\in E_{i,i}$ times the number of $j$ such
that $\gamma \derivrg{\ast}{G} a_{i+1} \cdots a_j$. 
The number of such $j$ is bounded by
$n+1$, and the number of $[A\de\mydot\gamma]\in E_{i,i}$ is bounded by 
${\cal E}_1$ plus $\size{\{\alpha\ |\ (S \de \alpha) \in P\}}$.
Therefore
we have
${\cal V}_4 \leq (n+1) \cdot 
({\cal E}_1 + \size{\{\alpha\ |\ (S \de \alpha) \in P\}})$

Steps~5 and~6 cannot be applied more than once for each application
of Steps~2 and~3 and $[\beta] \in T_{j,m}$, for at most $n+1$
different values of $m$.
Therefore
we have
${\cal V}_5\leq (n+1) \cdot {\cal V}_2\leq
(n+1) \cdot {\cal E}_2$ and ${\cal V}_6\leq (n+1) \cdot {\cal V}_3\leq
(n+1) \cdot {\cal E}_3$.

Combining the above, we obtain:
\begin{proposition}
For fixed \cfg\ and input of length $n$, we have
${\cal V} \leq (n+2) \cdot {\cal E}$.
\end{proposition}
In the worst case, the number of steps for the variant may thus be 
greater than the number of steps for the original Earley algorithm by a factor
which is $\order{n}$. 
Again, the 
data presented by the next section suggest that this consideration 
has little bearing on practical cases.

\section{Empirical Results}
\label{s:empiric}

We have performed some experiments with Algorithms~\ref{a:earley} 
and~\ref{a:variant} 
for four practical context-free grammars.

The first grammar
generates a subset of the programming language\linebreak ALGOL~68
\cite{vW75}. The second and third
grammars generate fragments of Dutch, and are
referred to
as the CORRie grammar \cite{VO94} and the Deltra
grammar \cite{SH90},
respectively.
These grammars were stripped of their arguments
in order to
convert them into context-free grammars.
The fourth grammar, referred to as the
Alvey grammar \cite{CA93},
generates a fragment of English and was
automatically generated from a unification-based grammar.

The test sentences
have been obtained by automatic generation from the grammars,
using a random
generator to select productions,
as explained in \cite{NE96f};
therefore these sentences do
not necessarily represent input typical
of the applications
for which the grammars were written.
Table~\ref{grammars} summarizes the test material.
\begin{table}[t]
\begin{center}
\renewcommand{\arraystretch}{1.1}
\begin{tabular}{|l||r|r|r||r|rl|}
\hline
$G = (\myterm,\nonterm,P,S)$ &
\multicolumn{1}{c|}{$|G|$} &
\multicolumn{1}{c|}{$|\nonterm|$}&
\multicolumn{1}{c||}{$|P|$} &
\multicolumn{1}{c|}{$|w|$} &
\multicolumn{2}{c|}{Parses}\\
\hline
\hline
ALGOL 68 & 783 & 167 & 330 & 13.7 & 2.6 * & $10^0$ \\
CORRie   & 1141 & 203 & 424 & 12.3 & 2.3 * & $10^{14}$  \\
Deltra   & 1929 & 281 & 703 & 10.8  & 1.1 * & $10^{73}$\\
Alvey    & 5072 & 265 & 1484 & 10.7  & 3.2 * & $10^4$  \\
\hline
\end{tabular}
\end{center}
\vspace{-1em}
\caption{The test material: the four grammars and some of their
dimensions, the average length of
the test sentences (20 sentences of various lengths for each grammar), and the
average number of parses per sentence 
(excluding parses containing cycles, i.e.\ subderivations of the form
$A\derivrg{+}{G} A$).}
\label{grammars}
\begin{center}
\renewcommand{\arraystretch}{1.0}
\begin{tabular}{|l||r|r||r|r|r|r||r|r|}
\hline
    & \multicolumn{2}{c||}{Earley}  & \multicolumn{4}{|c||}{Variant}
   & \multicolumn{2}{c|}{$\tau_2$ + Earley} \\
\cline{2-9}
$G$ &  \multicolumn{1}{c|}{${\cal E}$} &
       \multicolumn{1}{c||}{$|E|$} &
       \multicolumn{1}{c|}{${\cal V}$} &
       \multicolumn{1}{c|}{$|U|$} &
       \multicolumn{1}{c|}{$|T|$} &
       \multicolumn{1}{c||}{$|U|$+$|T|$} &
       \multicolumn{1}{c|}{${\cal E}$} &
       \multicolumn{1}{c|}{$|E|$}
\\
\hline
\hline
ALGOL 68 & 2,062 & 1,437 & 2,054 & 1,302 & 119 & 1,421 & 2,107 & 1,483 \\
CORRie & 19,164 & 8,361 & 15,492 & 3,498 & 2,746 & 6,244 & 17,450 & 8,751 \\
Deltra & 60,849 & 12,694 & 34,238 & 4,759 & 4,071 & 8,830 & 57,582 & 15,114 \\
Alvey & 47,562 & 6,304 & 27,786 & 5,398 & 180 & 5,578 & 47,552 & 6,314 \\
\hline
\end{tabular}
\end{center}
\vspace{-1em}
\caption{Dynamic requirements: average time and space per sentence.}
\vspace{-1.5em}
\label{results}
\end{table}

Our implementation is merely a prototype, which means that
absolute duration of the parsing process is little indicative of
the actual efficiency of more sophisticated implementations.
Therefore, our measurements have been restricted to implementation-independent
quantities, viz.\ the number of elements stored
in the parse table and the number of elementary steps
performed by the algorithm. In a
practical implementation, such quantities will strongly influence the
space and time complexity, although they do not represent the only
determining factors. Furthermore, all optimizations
of the time and space efficiency have been left out of consideration.

In our experiments we have also considered an alternative way of 
introducing suffix items $[\beta]$ (albeit only those
with $\size{\beta} \geq 2$)
into the parsing process, namely by first applying a grammar
transformation $\tau_2$, 
and then executing Algorithm~\ref{a:earley} as usual. 
This was motivated by the literature on 
{\em covers\/}~\cite{Nijholt:80,Leermakers:89AC}, 
which shows that some
complicated parsing algorithms can be simulated by means of 
grammar transformations and
simpler parsing algorithms. We have not found any way to completely simulate
Algorithm~\ref{a:variant} in this manner, 
but the following transformation captures
some of its behaviour.%
\footnote{Algorithm~\ref{a:variant} avoids any use of items of the form
$[A\de X \mydot Y ]$. The same cannot be achieved by means 
of a grammar transformation and
Algorithm~\ref{a:earley}. An alternative would be to apply some other kind of 
tabular algorithm to the transformed grammar. See e.g.\ \cite{NE96}.}
For an arbitrary grammar 
$G = (\myterm, \nonterm, P, S)$, we define $\tau_2(G) =
(\myterm, \nonterm\cup\ieset, P', S)$, where $P'$ contains the
following productions:
\\[.5ex]
\begin{tabular}{lll}
&$A\de X [\alpha]$ & for all $(A\de X \alpha)\in P$ 
                     with $\size{\alpha} > 1$; \\
\ \ \ &$A\de \alpha$ & for all $(A\de \alpha)\in P$ 
                     with $\size{\alpha} \leq 2$; \\
&$[X\alpha]\de X [\alpha]$\ \ \ \ & for all $[X\alpha] \in \ieset$ 
                     with $\size{\alpha} > 1$;  \\
&$[XY]\de X Y$ & for all $[XY] \in \ieset$. \\
\end{tabular}
\\[.5ex]
Note that the transformed grammar is in {\em two normal form},
which means that the length of right-hand sides of productions 
is at most~2.

Table~\ref{results} presents the costs of parsing the
test sentences.
These data show that 
there is a significant gain in
space and time efficiency in moving from Algorithm~\ref{a:earley} 
to Algorithm~\ref{a:variant}. 
The biggest improvement in the number of parsing
steps is observed in the case of the Alvey grammar, where it amounts
to a decrease by over 41\%. The biggest improvement in the total 
number of items stored in the tables occurs for de Deltra grammar, where
it amounts to a decrease by over 30\%. Only for {\em individual\/} 
sentences for ALGOL~68 was there an increase in time and space, by at most 
1.2\% and 0.2\%, respectively.

In the case of ALGOL 68 and Alvey, it is striking that
$T$ is so much smaller than $U$ and $E$. 
This may be explained by the relatively
low level of ambiguity, as compared to the other two grammars (see
Figure~\ref{grammars}).
Both the Earley algorithm and its variant predict many productions
in the form of items in $U$ and $E$, but
only a limited number of these 
productions will be recognized in their entirety,
resulting in items in $T$.
Although less striking in these cases, we see that
also for CORRie and Deltra $T$ is smaller than $U$.
This suggests that the potential undesirable behaviour of the variant
with regard to the original Earley algorithm, as discussed in the
previous section, does not occur in practice.

The approach using the grammar transformation is not competitive with the other two
approaches. Although the number of 
steps is sometimes slightly smaller than in the case of Algorithm~\ref{a:earley}, 
the space requirements are larger in all cases.

\section{Concluding Remarks}
\label{s:disc}

We have presented a variant of the Earley algorithm
and have discussed cases in which it achieves space 
and time savings with respect to the original algorithm. 
Our variant is based on the following two main ideas. 
First, we do not compute left positions of productions
until we are granted that production recognition can be completed
within the input.  Second, 
we only use suffix items as defined in~(\ref{e:ieset}). 

The idea of dropping left positions of productions has first 
been proposed by~\cite{LE92}, where a functional 
realization of left-corner parsing is presented. 
This idea was rediscovered by~\cite{DO94} and expressed 
in a more direct way, using a table similar to our table $U$.

The idea of using suffix items has also been proposed 
in~\cite{LE92}.  It has later been rediscovered by~\cite{DO94}. 
It was also applied to LR parsing in \cite{NE96}.
In the literature on chart parsing, e.g. in~\cite{BE83}, 
one sometimes also finds a weaker form
of this idea, where the set of items used in labeling edges is
$\chartset = \{ [A \de \beta] \sep (A \de \alpha \beta) \in P \}$.
One observes that, with respect to items $[A \de \alpha\mydot\beta]$ 
from $\eset$, the $\alpha$ is omitted as in the case of $\ieset$, 
yet the left-hand side $A$ is retained.
If this idea is not combined with the idea of dropping left positions,
then the benefit of this is limited to grammars containing many
pairs of productions of the form $A\de \alpha \beta$ and $A\de \gamma \beta$,
with $\alpha\neq\gamma$.
The idea of using suffix items is 
related to the difference between two kinds of Earley parsing for
the ID/LP formalism: in~\cite{SH84} the items are of the form
$[A\de\alpha\mydot\beta]$, where $\alpha$ is a string of constituents and
$\beta$ is a set of constituent, whereas in~\cite{BA85}, both $\alpha$
and $\beta$ are sets.
This allows representation of several items according to~\cite{SH84} 
by a single item according to~\cite{BA85}, as has been argued 
in~\cite[Section~9.2]{NA94}.

The ideas above rely on productions or items having some suffix in common.
Alternatively, one can investigate optimizations that rely on productions that have
{\em prefixes\/} in common~\cite{Nederhof:94}.

\bibliographystyle{plain}

\begin{thebibliography}{10}

\bibitem{BA85}
G.~E. Barton, Jr.
\newblock On the complexity of {ID/LP} parsing.
\newblock {\em Computational Linguistics}, 11(4):205--218, 1985.

\bibitem{BE83}
J.~Bear.
\newblock A breadth-first parsing model.
\newblock In {\em Proc.\ of the Eighth International Joint Conference on
  Artificial Intelligence}, volume~2, pages 696--698, Karlsruhe, West Germany,
  August 1983.

\bibitem{Billot:89}
S.~Billot and B.~Lang.
\newblock The structure of shared forests in ambiguous parsing.
\newblock In {\em Proc.\ of the 27$^{\it th}$ ACL}, pages 143--151, Vancouver,
  British Columbia, Canada, 1989.

\bibitem{CA93}
J.~A. Carroll.
\newblock Practical unification-based parsing of natural language.
\newblock Technical Report No. 314, University of Cambridge, Computer
  Laboratory, England, 1993.
\newblock PhD thesis.

\bibitem{DO94}
J.~Dowding, R.~Moore, F.~Andry, and D.~Moran.
\newblock Interleaving syntax and semantics in an efficient bottom-up parser.
\newblock In {\em Proc.\ of the 32$^{\it nd}$ ACL}, pages 110--116, Las Cruces,
  New Mexico, 1994.

\bibitem{Earley:70}
J.~Earley.
\newblock An efficient context-free parsing algorithm.
\newblock {\em Communications of the Association for Computing Machinery},
  13(2):94--102, 1970.

\bibitem{Gardemann:89}
D.~Gardemann.
\newblock Using restriction to optimize unification parsing.
\newblock In {\em International Workshop on Parsing Technologies}, pages 8--17,
  Pittsburgh, 1989.

\bibitem{Graham:76}
S.~L. Graham and M.~A. Harrison.
\newblock Parsing of general context free languages.
\newblock In {\em Advances in Computers}, volume~14, pages 77--185. Academic
  Press, New York, NY, 1976.

\bibitem{Graham:80}
S.~L. Graham, M.~A. Harrison, and W.~L. Ruzzo.
\newblock An improved context-free recognizer.
\newblock {\em {ACM} Transactions on Programming Languages and Systems},
  2(3):415--462, 1980.

\bibitem{Kaplan:73}
R.~Kaplan.
\newblock A general syntactic processor.
\newblock In E.~Rustin, editor, {\em Natural Language Processing}.
  Prentice-{Hall}, Englewood Cliffs, NJ, 1973.

\bibitem{Kay:86a}
M.~Kay.
\newblock Algorithm schemata and data structures in syntactic processing.
\newblock Technical report CSL-80, Xerox Palo Alto Research Center, Palo Alto,
  CA, 1980.
\newblock Also in: B. J. Grosz, K. {Sparck Jones} and B. L. Webber, editors,
  {\em Natural Language Processing}, pages 35-70, Kaufmann, Los Altos, CA,
  1986.

\bibitem{Leermakers:89AC}
R.~Leermakers.
\newblock How to cover a grammar.
\newblock In {\em Proc.\ of the 27$^{\it th}$ ACL}, pages 135--142, Vancouver,
  British Columbia, Canada, 1989.

\bibitem{LE92}
R.~Leermakers.
\newblock A recursive ascent {E}arley parser.
\newblock {\em Information Processing Letters}, 41(2):87--91, February 1992.

\bibitem{Leermakers:92}
R.~Leermakers.
\newblock Recursive ascent parsing: from {E}arley to {M}arcus.
\newblock {\em Theoretical Computer Science}, 104:299--312, 1992.

\bibitem{Leiss:90}
H.~Leiss.
\newblock On {K}ilbury's modification of {E}arley's algorithm.
\newblock {\em {ACM} Transactions on Programming Languages and Systems},
  12(4):610--640, 1990.

\bibitem{LE91b}
J.~M. I.~M. Leo.
\newblock A general context-free parsing algorithm running in linear time on
  every {LR($k$)} grammar without using lookahead.
\newblock {\em Theoretical Computer Science}, 82:165--176, 1991.

\bibitem{NA94}
S.~Naumann and H.~Langer.
\newblock {\em Parsing}.
\newblock B.G. Teubner, Stuttgart, 1994.

\bibitem{Nederhof:94}
M.~J. Nederhof.
\newblock An optimal tabular parsing algorithm.
\newblock In {\em Proc.\ of the 32$^{\it nd}$ ACL}, pages 117--124, Las Cruces,
  New Mexico, 1994.

\bibitem{NE96f}
M.~J. Nederhof.
\newblock Efficient generation of random sentences.
\newblock {\em Natural Language Engineering}, 2(1):1--13, 1996.

\bibitem{NE96}
M.~J. Nederhof and G.~Satta.
\newblock Efficient tabular {LR} parsing.
\newblock In {\em Proc.\ of the 34$^{\it th}$ ACL}, pages 239--246, Santa Cruz,
  CA, 1996.

\bibitem{Nijholt:80}
A.~Nijholt.
\newblock {\em Context-Free Grammars: Covers, Normal Forms, and Parsing},
  volume~93.
\newblock Springer-{Verlag}, Berlin, Germany, 1980.

\bibitem{PE83}
F.~C.~N. Pereira and D.~H.~D. Warren.
\newblock Parsing as deduction.
\newblock In {\em Proc.\ of the 21$^{\it st}$ ACL}, pages 137--144, Cambridge,
  MA, 1983.

\bibitem{SH90}
J.~J. Schoorl and S.~Belder.
\newblock Computational linguistics at {D}elft: A status report.
\newblock Report {WTM/TT} 90-09, Delft University of Technology, Applied
  Linguistics Unit, 1990.

\bibitem{Shann:91}
P.~Shann.
\newblock Experiments with {GLR} and chart parsing.
\newblock In M.~Tomita, editor, {\em Generalized {LR} Parsing}. Kluwer Academic
  Publishers, 1991.

\bibitem{SH84}
S.~M. Shieber.
\newblock Direct parsing of {ID/LP} grammars.
\newblock {\em Linguistics and Philosophy}, 7:135--154, 1984.

\bibitem{Shieber85AC}
S.~M. Shieber.
\newblock Using restriction to extend parsing algorithms for
  complex-feature-based formalisms.
\newblock In {\em Proc.\ of the 23$^{\it rd}$ ACL}, pages 145--152, Chicago,
  IL, 1985.

\bibitem{Stock:89CL}
O.~Stock.
\newblock Parsing with flexibility, dynamic strategies, and idioms in mind.
\newblock {\em Computational Linguistics}, 15(1):1--18, 1989.

\bibitem{vW75}
A.~van Wijngaarden et~al.
\newblock Revised report on the algorithmic language {ALGOL} 68.
\newblock {\em Acta Informatica}, 5:1--236, 1975.

\bibitem{VO94}
T.~G. Vosse.
\newblock {\em The Word Connection}.
\newblock PhD thesis, University of Leiden, 1994.

\bibitem{Wiren:87EA}
M.~Wiren.
\newblock A comparison of rule-invocation strategies in 
  parsing.
\newblock In {\em Proc.\ of the 3$^{\it rd}$ EACL}, pages 226--233, Copenhagen,
  Denmark, 1987.

\end{thebibliography}

\end{document}